\documentclass[runningheads]{llncs}
\usepackage{graphicx}
\usepackage{amssymb}
\usepackage{booktabs}
\usepackage{listings}
\usepackage{csquotes}
\usepackage{hyperref}

\begin{document}
	\title{Data-driven Real-time Short-term Prediction of Air Quality: Comparison of ES, ARIMA, and LSTM \thanks{Preprint}}
	\titlerunning{Short-term Prediction of Air Quality}
	%
	\author{	
       	Iryna~Talamanova\inst{1}    
         \and
		Sabri~Pllana\inst{2} 
	}
	\authorrunning{I. Talamanova and S. Pllana}
	%

	\institute{
            Stockholm, Sweden \\ 
            \email{iryna.talmanova@gmail.com} 
        \and
            Center for Smart Computing Continuum, Forschung Burgenland, Eisenstadt, Austria \\ 
            \email{sabri.pllana@forschung-burgenland.at} 
        }
	\maketitle              

\begin{abstract}
Air pollution is a worldwide issue that affects the lives of many people in urban areas. It is considered that the air pollution may lead to heart and lung diseases. A careful and timely forecast of the air quality could help to reduce the exposure risk for affected people. 
In this paper, we use a data-driven approach to predict air quality based on historical data. We compare three popular methods for time series prediction: Exponential Smoothing (ES), Auto-Regressive Integrated Moving Average (ARIMA) and Long short-term memory (LSTM). Considering prediction accuracy and time complexity, our experiments reveal that for short-term air pollution prediction ES performs better than ARIMA and LSTM.

\end{abstract}

\section{Introduction}
\label{sec:introduction}

The high-density of urban population may often lead to high-levels of various kinds of pollution, such as, air pollution  \cite{Yang:2022}, light pollution \cite{Longcore:2004}, or noise pollution\cite{Alsouda:2018,Alsouda:2019}. Air pollution is an important indicator of the life quality of a city. It may affect human health by causing heart and lung diseases~\cite{Das:2022}. Furthermore, air pollution is a significant contributor to climate change~\cite{Seinfeld:2016}. Table \ref{table:air_pollution_effects} describes the health impact of some of the most harmful air pollutants.

\begin{table}[ht]
\begin{center}
\caption{Impact of air pollution on human health.}
\label{table:air_pollution_effects}
\begin{tabular} {|c p{10cm}|}
\hline
\textbf{Pollutant} &  \textbf{Health Impact}  \\ 
\hline
CO         &  Mostly affects the cardiovascular system and leads to a lack of oxygen in the human body. The outcome is bad concentration and slow reflexes. It also may cause lung inflammation.\\

O3         &  Affects lungs and respiratory system and may lead to lung inflammation. It also reduces lung functions and may cause asthma or even lung cancer.\\

NO2, SO2   & Affects the respiratory system by reducing its resistance to respiratory infections. \\ 

PM         &  May cause inflammatory lung changes. It is dangerous due to its small size, which allows it to reach the heart and the brain and may cause inflammation there. \\ 
\hline
\end{tabular}
\end{center}
\end{table}

Sensors for personal use enable to measure the air pollution at a specific location. Based on the retrieved data, it is possible to predict future air pollution using time series forecasting methods. Existing solutions, are already capable to quantify the air quality state. However, these solutions do not provide short-term forecasts in real time.

Prediction of time series in near real time implies that the model should be periodically updated to account for recently generated data. Amjad et al. \cite{Amjad:2017} propose to monitor the trend of the financial time series and update the model if the trend changes direction. Qin et al. \cite{Qin:2004} suggest to create a schedule for updating the data within fixed time frame. For an adaptive model selection, Le et al. \cite{Le:2007} suggest to update the model each time new data arrives. A trend in the series of air pollution data may change unpredictably because of environmental changes. Therefore, updating the model each time new data arrives is a promising approach for air pollution data.

In this paper, we compare methods that may be applied to real-time prediction of air quality. We conduct a performance comparison with respect to the prediction accuracy and time complexity for three methods: Exponential Smoothing, ARIMA, and LSTM. For experimental evaluation we use a data set \cite{dataset-skopje} that comprises air pollution data collected in Skopje. Our experimental results indicate that Exponential Smoothing outperforms ARIMA and LSTM for real time prediction of air quality. 

The rest of the paper is structured as follows. We describe the data set, code implementation, and metrics in Section \ref{sec:implementation}. Section \ref{sec:eval} describes the experimental evaluation. Related work is discussed in Section \ref{sec:rw}. Section \ref{sec:conclusion} concludes the paper. 

\section{Methodology} 
\label{sec:implementation}

In this section, we describe data preparation, code implementation of prediction methods, and metrics. 

\subsection{Data preparation}

We use for experimental evaluation the data collected in Skopje \cite{dataset-skopje}. This data set contains air pollution measurements from seven stations, and includes CO, NO2, O3, PM10, and PM2.5. The pollutants behave quite similar over time, and our focus in this study is on PM2.5.

\begin{figure}[ht]
\begin{center}
\includegraphics[width=0.8\columnwidth]{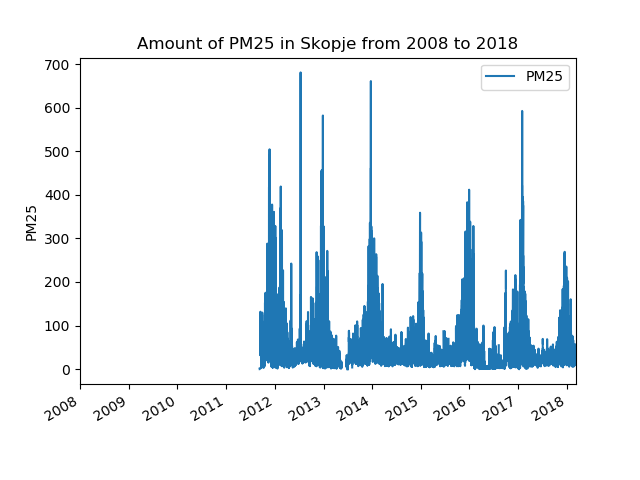}
\end{center}
\caption{Initial data set.}
\label{fig:initial_dataset}
\end{figure}

Figure \ref{fig:initial_dataset} visualizes the data set that originally had missing values and outliers.

\begin{figure}[ht]
\begin{center}
\includegraphics[width=0.8\columnwidth]{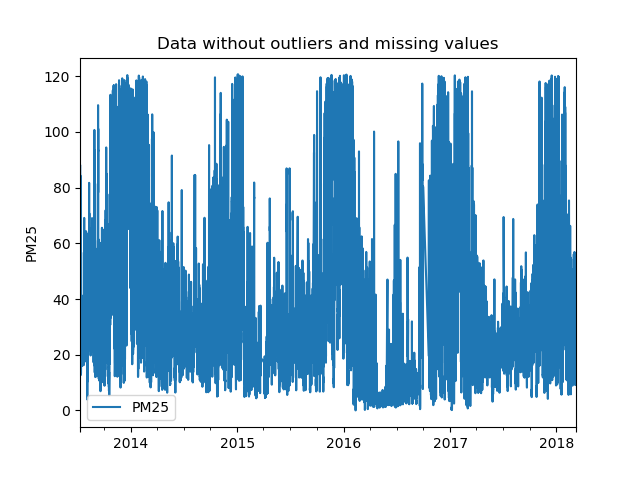}
\end{center}
\caption{Data set after filling missing values.}
\label{fig:no_missing_values}
\end{figure}

We use linear interpolation for filling missing values. It is assumed that missing values lie on the line which could be drawn using a set of known points. The data set after applying interpolation is depicted in Figure \ref{fig:no_missing_values}.

Because we are interested in short-term predictions, we decided to predict the air pollution 24 hours ahead; that is, the test interval is 24 hours. In accordance with \cite{Siami:2018} the training and test interval are split as 70\% and 30\% respectively. We should consider that the prediction accuracy of a Neural Network usually improves with the increase of the amount of training data. Therefore, for the LSTM model the training set is much larger than for statistical models.

\subsection{Code implementation}

We have implemented all algorithms in this study in Python using open source libraries and frameworks (including, statsmodels \cite{Massaron:2016}, pmdarima \cite{Pmdarima:2017}, keras \cite{Gulli:2017}, tensorflow \cite{Abadi:2016}). 

\subsection{Measurement metrics}

For the real-time environment emulation, we use the rolling window technique. After building the model and making the forecast, the data is moved one hour ahead and model building and forecasting is repeated.

\par
\begin{figure}[ht]
\begin{center}
\includegraphics[width=0.9\columnwidth]{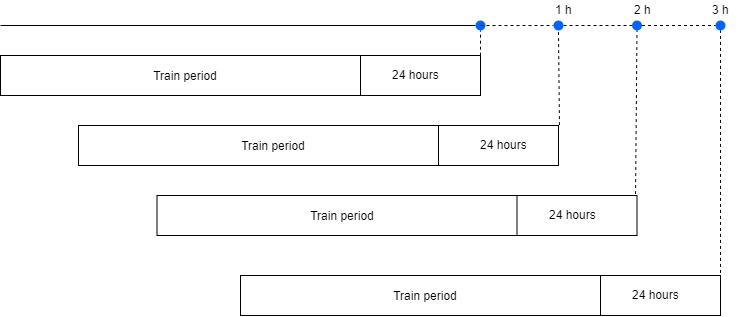}
\end{center}
\caption{Rolling forecast enables to emulate the real-time environment.}
\label{fig:rolling_forecast}
\end{figure}

 Figure \ref{fig:rolling_forecast} depicts the \emph{rolling forecast}. The forecast is done multiple times using different data and the error is calculated by averaging all errors that we have got. This technique is independent of the data and provides correct results.

We use the Root Mean Squared Error (RMSE), because it is considered to be less sensitive with respect to outliers. RMSE is the square root of the average of squared differences between the predicted value and actual value of the series.

\begin{equation}
RMSE=\sqrt{\frac{\sum_{t=1}^{n} (y_{i} -x_{i})^{2}}{n}}
\end{equation}

All experiments were conducted multiple times on different time intervals and the mean of the results has been taken. Also for each method, the multi-step forecast has been produced for comparison.

\section{Evaluation}
\label{sec:eval}

In this section, we present the training intervals and the Root Mean Squared Error (RMSE) for Exponential Smoothing, ARIMA, and LSTM. Additionally, a performance comparison of the studied methods is provided. 

\subsection{Exponential smoothing (ES)} 

To determine the best training intervals we conducted the following experiment. We have trained the model for various numbers of hours and selected the one that resulted with the lowest RMSE. Because we are aiming for short-term predictions, it is not necessary to consider larger training intervals.

Table \ref{table:table_es_train_selection} presents the average RMSE for different training intervals. We may observe that the best training interval for Exponential Smoothing is 96 hours, because RMSE for this period is the lowest. 

\begin{table}[ht]
\centering
\caption{RMSE of ES model for various training intervals.}
\label{table:table_es_train_selection}
\begin{tabular}{|c|c|c|c|c|c|c|c|}
\hline
Time [h]  & 48 & 72 & \textbf{96} & 120 & 144 & 168 & 196  
\\ \hline
RMSE  & 10.54 & 9.76 & \textbf{6.39} & 8.26 & 10.47 & 12.6 & 10.12  
\\ \hline
\end{tabular}
\end{table}

\subsection{ARIMA}

To make the prediction process faster and reduce the model selection time an experiment for determining hyperparameter intervals has been conducted. We have evaluated various time intervals to determine what the minimum and maximum values of the hyperparameters could be. The results of this experiment are presented in Table \ref{table:table_arima_parameters_selection}.

\begin{table}[ht]
\centering
\caption{ARIMA hyperparameter intervals.}
\label{table:table_arima_parameters_selection}
\begin{tabular}{|c|c|c|}
\hline
Hyperparameter & Minimum & Maximum \\ 
\hline
p & 0 & 6 \\ 
d & 0 & 2 \\ 
q & 0 & 5 \\ 
P & 0 & 3 \\ 
D & 0 & 1 \\ 
Q & 0 & 2 \\ 
\hline
\end{tabular}
\end{table}

After the hyperparameters boundaries were determined, we observed a significant improvement in the speed of building an ARIMA model. 

Table \ref{table:table_arima_train_selection} indicates that the best training interval for the ARIMA model is 120 hours, because it results with the lowest RMSE across the considered intervals. 

\begin{table}[ht]
\centering
\caption{RMSE of ARIMA model for various training intervals}
\label{table:table_arima_train_selection}
\begin{tabular}{|c|c|c|c|c|c|c|c|}
\hline
Time [h] & 72 & 96 & \textbf{120} & 144 & 168 & 196 & 220
\\ 
\hline
RMSE & 11.24 & 8.89 & \textbf{8.59} & 12.1 & 14.32 & 12.83 & 11.56   \\ 
\hline
\end{tabular}
\end{table}

The time for building the ARIMA model and performing prediction is 19.8 seconds in our experiment. While the time required for building ARIMA model and performing prediction is shorter than for Exponential Smoothing (20.5 seconds), the prediction error of ARIMA model is larger.

\subsection{LSTM}

We have implemented LSTM using Keras framework \cite{Gulli:2017}. In the first stage of the experiment, we evaluated different LSTM configurations: Simple, Stacked, Bidirectional, and Encoder-decoder. 

\begin{table}[ht]
\centering
\caption{Accuracy and execution time of various LSTM network configurations.}
\label{table:LSTM_network_configuration}
\begin{tabular}{|l|c|c|}
\hline
\textbf{Network Configuration} & \textbf{RMSE} & \textbf{Time [s]} \\
\hline
Simple                &  \textbf{3.26} & \textbf{1196.32} \\
Stacked               &  3.32 & 1253.21 \\
Bidirectional         &  4.19 & 1372.65 \\ 
Encoder-decoder       &  3.75 & 1476.87 \\
\hline
\end{tabular}
\end{table}

Evaluation results of LSTM network configurations are presented in Table \ref{table:LSTM_network_configuration}. We may observe that the best accuracy is achieved using a Simple LSTM configuration. In what follows in this section, we describe experimental results for the Simple LSTM configuration. Table \ref{table:lstm_results_hyperparameters_tuning} shows the selected values of hyperparameters for the Simple LSTM configuration.

\begin{table}[ht]
\centering
\caption{LSTM hyperparameter values.}
\label{table:lstm_results_hyperparameters_tuning}
\begin{tabular}{|l|c|}
\hline
\textbf{Hyper-parameter} & \textbf{Value} \\
\hline
Epoch & 800 \\ 
Patience coefficient & 0.1 \\ 
Validation size & 72 hours \\ 
Dropout & 0.1 \\ 
Recurrent dropout & 0.3 \\ 
Batch size & 12 \\ 
Type & Statefull \\ 
Coefficient for counting units & 3 \\ 
Training size & 8000 \\
\hline
\end{tabular}
\end{table}

Table \ref{table:table_rmse_lstm} shows RMSE and execution time of the Simple LSTM configuration for various values of prediction horizon. 

We may observe that the execution time of LSTM is significantly higher than for the other methods considered in this study (that is, ES and ARIMA). Furthermore, the prediction accuracy worsens with the increase of the prediction horizon.

\begin{table}[ht]
\centering
\caption{RMSE and execution time of the Simple LSTM configuration for various values of Horizon.}
\label{table:table_rmse_lstm}
\begin{tabular}{|c|r|r|}
\hline
\textbf{Horizon [h]} & \textbf{RMSE} & \textbf{Time [s]}\\
\hline
 1 &  3.26 & 1282.46  \\ 
 2 &  4.47 & 953.40   \\ 
 3 &  5.35 & 1140.42  \\ 
 4 &  6.17 & 1185.93  \\ 
 5 &  7.07 & 1101.65  \\ 
 6 &  7.31 & 967.06   \\ 
 7 &  8.24 & 1303.88  \\ 
 8 &  6.21 & 900.61   \\ 
 9 &  9.06 & 1111.17  \\ 
10 &  9.14 & 1055.00  \\ 
11 &  9.67 & 1354.31  \\ 
12 & 10.41 & 1081.18  \\ 
13 & 10.44 & 1258.37  \\ 
14 & 11.06 & 1195.12  \\ 
15 & 10.71 & 936.59   \\ 
16 & 11.29 & 1107.04  \\ 
17 & 11.88 & 1213.67  \\ 
18 & 13.40 & 1155.57  \\ 
19 & 12.02 & 977.91   \\ 
20 & 12.55 & 966.46   \\ 
21 & 12.71 & 1199.05  \\ 
22 & 12.93 & 1078.39  \\ 
23 & 11.74 & 1003.03  \\ 
24 & 14.69 & 1117.89  \\ 
\hline
\end{tabular}
\end{table}

\subsection{Performance comparison of ES, ARIMA, and LSTM}

\begin{figure}[ht]
\begin{center}
\includegraphics[width=0.8\columnwidth]{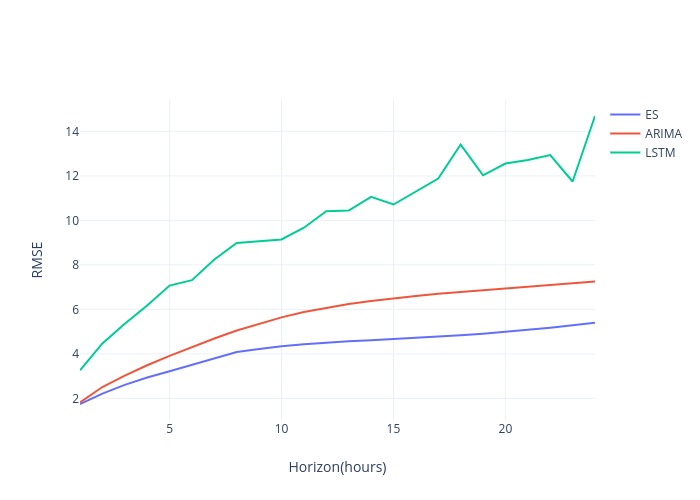}
\end{center}
\caption{RMSE and prediction Horizon [h] for ES, ARIMA, and LSTM.}
\label{fig:rmse_total}
\end{figure}

Figure \ref{fig:rmse_total} depicts the relationship of RMSE and prediction horizon for ES, ARIMA, and LSTM. RMSE increases for all three considered methods with the increase of prediction horizon. For all considered values of prediction horizon, ES has the lowest RMSE (that is, the highest prediction accuracy) compared to ARIMA and LSTM. 

With respect to the time for building a model and making a forecast, ARIMA and ES require about 20 seconds in our experiments. LSTM is significantly slower and requires about 1000 seconds for model building and forecasting.

Considering prediction accuracy (that is, RMSE) and model building and forecasting time, we may conclude that ES is the most suitable among the studied methods for short-term prediction of air pollution.

\subsection{Future research directions}

The future work may investigate the performance of additional prediction methods in the context of air pollution. Techniques for parallel processing (\cite{Amaral:2020,multicore_book}), acceleration (\cite{Viebke:2019,Memeti:2015,Benkner:2011}), and intelligent parameter selection (\cite{Memeti:2021}) could be studied to further improve the efficiency. 

\section{Related Work}
\label{sec:rw}

In this section, we provide an overview of the related work. 

Shaban et al. \cite{Shaban:2016} investigate three machine learning algorithms that could be used for alarming applications in the context of air pollution: Support Vector Machines,  M5P  Model  Trees, and  Artificial  Neural  Networks. In future they plan to study real-time prediction of air pollution.

Subramanian \cite{Subramanian:2016} studies application of Multiple Linear Regression and Neural Networks to forecasting of the pollution concentration. 

Le et al. \cite{Le:2018} describe an air-pollution prediction model that is based on spatio-temporal data that is collected from air quality sensors installed on taxis running across the city Daegu, Korea. The prediction model is based on the Convolutional Neural Network for an image like spatial distribution of air pollution. The temporal information in the data is handled using a combination of a Long Short-Term Memory unit for time series data and a Neural Network model for other air pollution impact factors (such as, weather conditions.)

Ochando et al. \cite{Ochando:2015} use traffic and weather data for prediction of the air pollution. The aim is to provide general information about the air quality of the city, and they do not focus on a particular spot within the city. In this study, the Random Forest model performs best.

Related work mostly focuses on studying air pollution data that has been already generated and stored in the past. In contrast to related work, we focus on using continuously updated data (that may be generated by sensors) for short-term prediction of the air pollution in near real-time. 

\section{Conclusions}
\label{sec:conclusion}

We have presented various methods that may be applied to short-term air quality prediction in real time. Using a real-world data set we conducted a performance comparison of three methods: Exponential Smoothing, ARIMA, and LSTM.

We have observed that Exponential Smoothing performs more efficiently for the short-term air pollution prediction compared to ARIMA and LSTM. LSTM has larger prediction error and takes more time to make the prediction. For the multi-step ahead forecast the gap between prediction errors of these methods grows with the increase of the number of steps. 

%


\begin{thebibliography}{10}
\providecommand{\url}[1]{\texttt{#1}}
\providecommand{\urlprefix}{URL }
\providecommand{\doi}[1]{https://doi.org/#1}

\bibitem{Abadi:2016}
Abadi, M., Barham, P., Chen, J., Chen, Z., Davis, A., Dean, J., Devin, M.,
  Ghemawat, S., Irving, G., Isard, M., et~al.: Tensorflow: A system for
  large-scale machine learning. In: 12th $\{$USENIX$\}$ Symposium on Operating
  Systems Design and Implementation ($\{$OSDI$\}$ 16). pp. 265--283 (2016)

\bibitem{Alsouda:2018}
Alsouda, Y., Pllana, S., Kurti, A.: {A Machine Learning Driven IoT Solution for
  Noise Classification in Smart Cities}. In: Machine Learning Driven
  Technologies and Architectures for Intelligent Internet of Things (ML-IoT).
  pp.~1--6. Euromicro (2018). \doi{10.48550/arXiv.1809.00238}

\bibitem{Alsouda:2019}
Alsouda, Y., Pllana, S., Kurti, A.: {IoT-based Urban Noise Identification Using
  Machine Learning: Performance of SVM, KNN, Bagging, and Random Forest}. In:
  Proceedings of the International Conference on Omni-Layer Intelligent
  Systems. pp. 62--67. COINS '19, ACM, New York, NY, USA (2019).
  \doi{10.1145/3312614.3312631}

\bibitem{Amaral:2020}
Amaral, V., Norberto, B., Goulão, M., Aldinucci, M., Benkner, S., Bracciali,
  A., Carreira, P., Celms, E., Correia, L., Grelck, C., Karatza, H., Kessler,
  C., Kilpatrick, P., Martiniano, H., Mavridis, I., Pllana, S., Respício, A.,
  Simão, J., Veiga, L., Visa, A.: Programming languages for data-intensive
  {HPC} applications: A systematic mapping study. Parallel Computing
  \textbf{91},  102584 (2020).
  \doi{https://doi.org/10.1016/j.parco.2019.102584}

\bibitem{Amjad:2017}
Amjad, M., Shah, D.: Trading bitcoin and online time series prediction. In:
  NIPS 2016 Time Series Workshop. pp. 1--15 (2017)

\bibitem{Benkner:2011}
Benkner, S., Pllana, S., Traff, J., Tsigas, P., Dolinsky, U., Augonnet, C.,
  Bachmayer, B., Kessler, C., Moloney, D., Osipov, V.: {PEPPHER: Efficient and
  Productive Usage of Hybrid Computing Systems}. IEEE Micro  \textbf{31}(5),
  28--41 (2011). \doi{10.1109/MM.2011.67}

\bibitem{Das:2022}
Das, P.K., A, D.V., Meher, S., Panda, R., Abraham, A.: A systematic review on
  recent advancements in deep and machine learning based detection and
  classification of acute lymphoblastic leukemia. IEEE Access  \textbf{10},
  81741--81763 (2022). \doi{10.1109/ACCESS.2022.3196037}

\bibitem{Gulli:2017}
Gulli, A., Pal, S.: Deep Learning with Keras. Packt Publishing Ltd (2017)

\bibitem{Le:2018}
Le, D.: Real-time air pollution prediction model based on spatiotemporal big
  data. arXiv preprint arXiv:1805.00432  (2018)

\bibitem{Le:2007}
Le~Borgne, Y.A., Santini, S., Bontempi, G.: Adaptive model selection for time
  series prediction in wireless sensor networks. Signal Processing
  \textbf{87}(12),  3010--3020 (2007)

\bibitem{Longcore:2004}
Longcore, T., Rich, C.: Ecological light pollution. Frontiers in Ecology and
  the Environment  \textbf{2}(4),  191--198 (2004)

\bibitem{Massaron:2016}
Massaron, L., Boschetti, A.: Regression Analysis with Python. Packt Publishing
  Ltd (2016)

\bibitem{Memeti:2015}
Memeti, S., Pllana, S.: {Accelerating DNA Sequence Analysis Using Intel(R) Xeon
  Phi(TM)}. In: 2015 IEEE Trustcom/BigDataSE/ISPA. vol.~3, pp. 222--227 (2015).
  \doi{10.1109/Trustcom.2015.636}

\bibitem{Memeti:2021}
Memeti, S., Pllana, S.: Optimization of heterogeneous systems with {AI}
  planning heuristics and machine learning: a performance and energy aware
  approach. Computing  \textbf{103}(12),  2943--2966 (Dec 2021).
  \doi{10.1007/s00607-021-01017-6}

\bibitem{Ochando:2015}
Ochando, L.C., Juli\'{a}n, C.I.F., Ochando, F.C., Ferri, C.: Airvlc: An
  application for real-time forecasting urban air pollution. In: Proceedings of
  the 2nd International Conference on Mining Urban Data - Volume 1392. p.
  72–79. MUD'15, CEUR-WS.org, Aachen, DE (2015)

\bibitem{dataset-skopje}
Petrushevski, S.: {Air pollution in Skopje from 2008 to 2018} (2018),
  \url{https://www.kaggle.com/cokastefan/pm10-pollution-data-in-skopje-from-2008-to-2018}

\bibitem{multicore_book}
Pllana, S., Xhafa, F.: {Programming Multicore and Many-core Computing Systems}.
  John Wiley \& Sons, Inc., Hoboken, New Jersey, USA, 1 edn. (2017).
  \doi{10.1002/9781119332015}

\bibitem{Qin:2004}
Qin, X., Mahmassani, H.S.: Adaptive calibration of dynamic speed-density
  relations for online network traffic estimation and prediction applications.
  Transportation research record  \textbf{1876}(1),  82--89 (2004)

\bibitem{Seinfeld:2016}
Seinfeld, J.H., Pandis, S.N.: Atmospheric chemistry and physics: from air
  pollution to climate change. John Wiley \& Sons (2016)

\bibitem{Shaban:2016}
Shaban, K.B., Kadri, A., Rezk, E.: Urban air pollution monitoring system with
  forecasting models. IEEE Sensors Journal  \textbf{16}(8),  2598--2606 (April
  2016). \doi{10.1109/JSEN.2016.2514378}

\bibitem{Siami:2018}
Siami-Namini, S., Namin, A.S.: Forecasting economics and financial time series:
  Arima vs. lstm. arXiv preprint arXiv:1803.06386  (2018)

\bibitem{Pmdarima:2017}
Smith, T.G.: {pmdarima}: Arima estimators for {Python} (2017)

\bibitem{Subramanian:2016}
Subramanian, V.N.: Data analysis for predicting air pollutant concentration in
  smart city uppsala (2016)

\bibitem{Viebke:2019}
Viebke, A., Memeti, S., Pllana, S., Abraham, A.: {CHAOS: a parallelization
  scheme for training convolutional neural networks on Intel Xeon Phi}. The
  Journal of Supercomputing  \textbf{75}(1),  197--227 (Jan 2019).
  \doi{10.1007/s11227-017-1994-x}

\bibitem{Yang:2022}
Yang, D., Wang, J., Yan, X., Liu, H.: Subway air quality modeling using
  improved deep learning framework. Process Safety and Environmental Protection
   \textbf{163},  487--497 (2022).
  \doi{https://doi.org/10.1016/j.psep.2022.05.055}

\end{thebibliography}
\end{document}